%% file: main.tex
\documentclass[10pt,twocolumn]{article}

\usepackage[preprint]{antintlpaper}

\usepackage{marvosym}
\usepackage{hyperref}
\usepackage{makecell}
\usepackage{multirow}
\usepackage{colortbl}
\usepackage{algorithm}
\usepackage{algorithmic}

\AntTitle{Diagnosis Before Recovery: Turning Agent Failures into Selective Self-Correction}
\AntAuthors{
Pan Wang \and
Yihao Hu \and
Hang Wang \and
Zirui Lv \and
Xin Zhang \and
Jianshe Li \and
Jiang-Ming Yang \\
Wei Wu \and
Yongqi Tong\Letter
}
\AntAffiliations{Ant International}
\AntContact{
Contact: Pan Wang \href{mailto:diogenescask@gmail.com}{diogenescask@gmail.com}

$\textrm{\Letter}$ Corresponding Author: 
\href{mailto:yoqitong@gmail.com}{yoqitong@gmail.com}
}
\AntDate{\today}
\newcommand{\circnum}[1]{\tikz[baseline=(char.south)]{\node[shape=circle,fill=black,text=white,inner sep=0pt,minimum size=0.8em,font=\scriptsize\rmfamily\bfseries] (char) {\scriptsize #1};}}
\newcommand{\markissue}[2]{\hypertarget{#1}{\circnum{#2}}}
\newcommand{\refissue}[2]{\hyperlink{#1}{\circnum{#2}}}

\AntKeywords{language agents, self-correction, failure diagnosis, recovery policies, agent evaluation}
\AntAbstract{%
Self-correction is particularly useful when a failure constrains the next repair. Coding agents benefit from this property because compilers, tests, and execution traces turn many failures into typed recovery signals, but broad language-agent tasks often expose only a coarse task failure. This creates a tension for generic recovery playbooks: they broaden the agent's context precisely when the system needs a narrower repair interface, mixing incompatible signals for invalid actions, missing procedures, and strict-format errors. Our insight is that development-set failures can recover part of the missing diagnostic substrate by deciding which recovery interventions are admissible before test-time correction. We propose DARC, a diagnosis-guided recovery harness that profiles task-family failure modes, prunes mismatched interventions from a shared recovery library, and freezes a verifier-selected success-cost policy for deployment. This causal order makes correction selective: the harness first determines what kind of failure can be repaired, then decides how much recovery evidence to spend. In ALFWorld, AppWorld, and XBRL Finance, the same protocol yields an action-validity harness, a procedural-recovery fallback, and a format-precision retrieval policy; in each evaluated setting it improves average task performance over base agents and broad playbooks while reducing environment steps or retrieval budget. Our experiments show that failures need not trigger uniformly more context: DARC turns self-correction from prompt expansion into recovery-interface design. DARC provides a practical route toward more reliable agents in domains where compiler-like feedback is absent: making failures actionable before making contexts larger.
}

\begin{document}

\makeanttitle

\input{sections/01-introduction}
\input{sections/02-related_work}

\input{sections/03-method}
\input{sections/04-experiments}
\input{sections/05-conclusion}

% \section*{Acknowledgements}
% For public technical reports, acknowledge collaborators, infrastructure, and
% funding as appropriate. Remove this section for anonymous submissions.

% \section*{Impact Statement}
% Describe the intended benefit, plausible misuse, affected stakeholders, and
% the safeguards or evaluation boundaries relevant to deployment.

\bibliographystyle{plainnat}
\bibliography{references}

\appendix
\input{sections/appendix}

\end{document}

%% file: sections/01-introduction.tex
\section{Introduction}
\label{sec:intro}

Coding agents have become one of the most visible successes of language-agent research. A major reason is not only that code models improved, but that software environments provide an unusually powerful recovery interface: compilers expose syntax and type errors, tests expose behavioral mismatches, and execution traces localize intermediate states~\cite{chen2023selfdebugging,zhong2024ldb,jimenez2024swebench,yang2024sweagent,li2026rethinking}. These signals do more than add information. They make failure actionable. A type error, a failing assertion, and a timeout call for different repairs; the environment helps the agent distinguish them.

Most agent tasks are asked to self-correct without this diagnostic substrate. An embodied household agent may fail after issuing an invalid action, but the benchmark may only report that the episode was not solved. A multi-application assistant may miss a goal because it lacks the right API workflow, while a financial extraction agent may locate the evidence but violate a strict schema. These are not the same failure. Yet many recovery recipes treat them as if they were: append reflection, retrieve more examples, expose a longer tool manual, or run a broad recovery playbook. That strategy is attractive because it is generic, but it suffers from three concrete problems. \markissue{issue:mismatch}{1} \textbf{Intervention Mismatch}: a recovery signal chosen without diagnosis often cannot address the failure at hand, such as exposing an exhaustive API manual to fix an invalid embodied action or a strict-format violation. \markissue{issue:interference}{2} \textbf{Recovery Interference}: irrelevant procedures can pull an embodied agent away from state-compatible actions, long manuals can mix incompatible workflow rules, and demonstrations for one failure mode can degrade another. \markissue{issue:cost}{3} \textbf{Uncontrolled Recovery Cost}: triggering every recovery signal on every failure inflates environment steps, retrieval budget, and inference tokens regardless of whether the extra evidence is needed.

The key question is therefore not whether agents need more recovery material, but \emph{which failure should be allowed to trigger which recovery signal}. This paper studies \emph{diagnosis-guided agent self-correction}: using development failures to recover part of the compiler-like role that broad agent tasks lack. Instead of treating a failure as a request for more context, we treat it as evidence about the dominant bottleneck of a task family. The output is a \emph{recovery harness}: a frozen policy that links a diagnosed failure mode to a restricted set of admissible recovery interventions, such as an action-validity guard, a procedural API source, a local induction rule, or a retrieval-demonstration budget.

We propose \textbf{DARC}, a framework for \textbf{D}iagnosis-guided \textbf{A}gent \textbf{R}ecovery and \textbf{C}orrection. DARC makes two design choices explicit. \textbf{First}, to address \refissue{issue:mismatch}{1}, it profiles development-set failures and restricts the recovery library before test evaluation, so interventions that cannot address the diagnosed failure mode are excluded rather than mixed into a generic playbook. \textbf{Second}, to resolve \refissue{issue:interference}{2} and \refissue{issue:cost}{3}, within the admissible set it uses training-set verifier feedback to distill a short success-cost policy for deployment, so that expensive recovery evidence is spent only when cheaper interventions fail. To ensure stability and auditability, we freeze this diagnosis at the task-family level rather than learning it online.

We evaluate this protocol in three settings chosen to stress different recovery needs. ALFWorld tests action validity, where recovery should constrain state-incompatible actions rather than expose unrelated manuals. AppWorld tests procedural breadth, where recovery should reveal richer API workflow knowledge only when cheaper procedural sources fail. XBRL Finance tests format precision, where recovery should tune demonstration evidence without expanding into a broad playbook. These settings allow us to evaluate whether failure diagnosis makes self-correction more reliable and cost-aware across diverse bottlenecks.

Our contributions are:
\begin{itemize}[leftmargin=*]
  \item We formulate diagnosis-guided agent self-correction as the problem of turning ambiguous task failures into actionable recovery signals, motivated by the compiler-like feedback available in coding but often absent in broad agent tasks.
  \item We propose DARC, which constructs recovery harnesses by development-set failure diagnosis, admissible-intervention restriction, and success-cost policy distillation.
  \item We evaluate DARC on ALFWorld, AppWorld, and XBRL Finance, showing that targeted recovery improves average success or accuracy under the reported protocols while reducing environment steps or retrieval budget.
  \item We provide extensive ablations isolating the impact of correct, generic, and mismatched recovery policies, demonstrating that diagnosis-guided restriction is essential for efficient self-correction.
\end{itemize}

\begin{figure*}[t]
\centering
\includegraphics[width=1.0\textwidth]{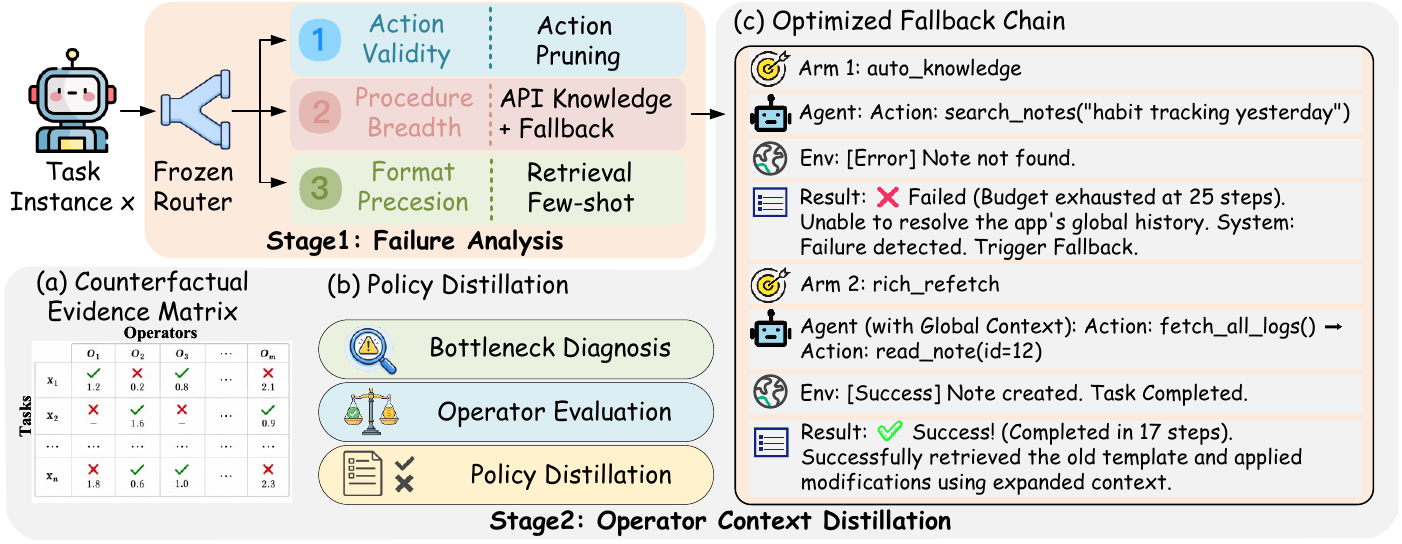}
\caption{DARC constructs a diagnosis-guided recovery harness. Development failures are converted into a task-family failure diagnosis, which restricts the candidate recovery interventions. Training-set verifier feedback then scores candidate fallback policies, and the selected policy is frozen before test evaluation.}
\label{fig:main_pipeline}
\end{figure*}

%% file: sections/02-related_work.tex
\section{Related Work}

\textbf{Self-correction from execution feedback.}
Recent work shows that agents can improve when feedback is structured enough to guide repair. In code generation, Self-Debugging uses execution results and explanations to refine programs~\cite{chen2023selfdebugging}, while LDB segments programs into execution units and verifies intermediate runtime states~\cite{zhong2024ldb}. Related code-agent and program-synthesis work uses generated tests, execution traces, benchmarks, or agent-computer interfaces to expose concrete repair signals~\cite{chen2022codet,ni2023lever,chen2021codex,austin2021program,jimenez2024swebench,yang2024sweagent,xia2024agentless,wang2024opendevin}. Work on learning from mistakes and negative examples similarly suggests that errors can be useful supervision when their structure is exposed~\cite{tong2024mistakes,wang2024learning,wang2026atlasva,hu2026seal}. Voyager uses environment feedback, execution errors, and self-verification to grow a skill library in Minecraft~\cite{wang2023voyager}. DARC takes this compiler-like feedback as motivation, but targets settings where such feedback is not naturally available: it uses development failures to build a task-family recovery interface for non-coding agent tasks.

\textbf{Language-agent recovery and reflection.}
ReAct~\cite{yao2022react} interleaves reasoning and acting, Reflexion~\cite{shinn2023reflexion} stores verbal feedback, and recursive critique~\cite{kim2023language} uses feedback loops for computer-control tasks. Broader reasoning and refinement methods use chain-of-thought, self-consistency, decomposition, search, or process supervision to make intermediate reasoning states more explicit~\cite{wei2022chain,kojima2022large,wang2023selfconsistency,zelikman2022star,zhou2023leasttomost,yao2023tree,madaan2023selfrefine,welleck2023selfcorrect,lightman2023process,cobbe2021gsm8k,tong2023iep,tong2024weak,wang2026reflex}. These methods demonstrate the value of self-correction, but the recovery mechanism is often applied uniformly across failures. DARC instead makes the recovery policy conditional on a diagnosed failure mode: action-validity recovery, procedural recovery, and format-precision recovery expose different interventions.

\textbf{Prompt optimization and generic context.}
Frameworks such as DSPy~\cite{khattab2023dspy} and teleprompter-based optimization~\cite{opsahl2024optimizing} tune instructions and demonstrations over validation data. Tool and multi-agent systems expose external modules, APIs, planning roles, or conversation protocols to language models~\cite{karpas2022mrkl,schick2023toolformer,paranjape2023art,li2023apibank,shen2023hugginggpt,qin2023toolllm,patil2024gorilla,wu2023autogen,li2023camel,hong2024metagpt,sumers2024cognitive}. Other lines use reinforcement learning or preference optimization to update model behavior~\cite{ouyang2022training,bai2022constitutional,lee2023rlaif,shao2024deepseekmath,rafailov2023direct,ahmadian2024back,li2023remax,wang2025bpo,xu2026rewardconsistency,pan2026selectiverm}, including multi-turn or device-control agents~\cite{zhou2024archer,bai2024digirl,qi2025webrl}. These approaches are complementary to DARC. Our focus is not weight updates or arbitrary prompt search, but the construction of a recovery interface that decides which feedback and interventions are relevant for a diagnosed failure mode.

\textbf{When more context is not neutral.}
The motivation for diagnosis-guided recovery is strengthened by evidence that irrelevant or poorly placed context can hurt language-model behavior. GSM-IC shows that irrelevant context can distract LLM reasoning~\cite{shi2023distracted}, while Lost in the Middle shows that models do not robustly use long contexts even when relevant information is present~\cite{liu2024lost}. Long-context, real-world chat, format-following, and safety studies similarly show that performance depends on where information appears, whether irrelevant material is present, and whether the task constraints are precise~\cite{bai2024longbench,hsieh2024ruler,li2024loogle,yang2025llm,tam2024let,lin2023toxicchat,pan2025overrefusal}. In agent settings, a generic playbook can similarly mix unrelated procedures, constraints, and demonstrations. DARC addresses this by restricting recovery context before policy search rather than appending a universal context bundle after every failure.

\textbf{Cascades, routing, and adaptive computation.}
LLM cascade methods such as FrugalGPT~\cite{chen2023frugalgpt} and AutoMix~\cite{aggarwal2023automix} route queries among models to trade cost for accuracy. Retrieval-augmented and adaptive-retrieval methods decide what evidence to fetch or how much retrieval is needed for a query~\cite{lewis2020retrieval,karpukhin2020dense,guu2020realm,izacard2021fid,izacard2022few,gao2023hyde,asai2024selfrag,yan2024crag,jeong2024adaptive}. DARC routes among \emph{recovery interventions} rather than base models. Scaling or switching the model does not necessarily fix an invalid action, a missing API procedure, or an output-format violation. The closest matched baseline is therefore a validation-selected full-library recovery cascade without diagnosis; we report this controlled comparison on ALFWorld in Section~\ref{sec:analysis} (Table~\ref{tab:diagnosis-isolation}), where diagnosis matches full-library accuracy while searching a much smaller policy space.

%% file: sections/03-method.tex
\section{Problem Setup: Diagnosis-Guided Recovery}
\label{sec:problem}

Let $\mathcal{X}$ denote a task family and let an agent produce an output or trajectory $y$ for each task $x \in \mathcal{X}$. Evaluation is given by a task verifier $s(x,y)\in\{0,1\}$ and an execution cost $c(x,y)\geq 0$, such as environment steps, latency, or retrieval budget. We assume a library of candidate recovery interventions $\mathcal{R}$, where an intervention may be an action guard, an API procedure source, a local induction rule, or a few-shot retrieval budget. We use \emph{intervention} for an executable correction signal that can be attached to a base agent and evaluated by the task verifier. A failure signal is \emph{actionable} when it selects or parameterizes an admissible intervention rather than merely adding more undifferentiated context.

The central object in this paper is a \emph{recovery harness}. A harness for failure mode $m$ is defined as a tuple
\begin{equation}
H_m = (m, \mathcal{R}_m, \pi_m),
\end{equation}
where $\mathcal{R}_m \subseteq \mathcal{R}$ is the set of interventions admissible for that diagnosed failure mode and $\pi_m=(r_1,\ldots,r_L)$ is the ordered deployment policy over those interventions. In our instantiation, $m$ is assigned at the task-family level from development-set failure profiles. Each benchmark family or split pair defines one $\mathcal{X}$ and one fixed recovery policy before test evaluation.

The recovery objective is to select a short policy that improves task success without invoking unnecessary or mismatched interventions:
\begin{equation}
\label{eq:problem_objective}
\max_{\pi \in \Sigma(\mathcal{R}_m)} \; \mathbb{E}_{x \sim \mathcal{X}}[s(x,\pi)] - \lambda \max\big(0,\mathbb{E}_{x \sim \mathcal{X}}[c(x,\pi)]-\tau_{\text{free}}\big),
\end{equation}
where $\Sigma(\mathcal{R}_m)$ is the space of bounded-length recovery policies, $\lambda$ controls the cost-success tradeoff, and $\tau_{\text{free}}$ is the cost allowance of a standard attempt. This formulation makes two questions explicit: which recovery signals are actionable for this failure mode, and in what order should they be invoked?

\section{DARC: Diagnosis-Guided Agent Recovery and Correction}

DARC constructs a recovery harness in two stages (Figure~\ref{fig:main_pipeline}). The diagnosis stage turns development failures into a task-family failure mode and restricts the recovery interventions that may be used. The policy-distillation stage freezes a cost-aware fallback policy over that restricted intervention set. Algorithm~\ref{alg:darc} summarizes the procedure.

\begin{algorithm}[t]
\caption{DARC Recovery-Harness Construction}
\label{alg:darc}

\begin{algorithmic}[1]

\REQUIRE Task family $\mathcal{X}$, candidate intervention library $\mathcal{R}$,
training tasks $\mathcal{D}_{\text{train}}$, failure-profile data
$\mathcal{D}_{\text{dev}}$, penalty $\lambda$.

\STATE \textbf{Profile:} Run the base agent on $\mathcal{D}_{\text{dev}}$
and measure failure signatures.

\STATE \textbf{Diagnose:} Select dominant failure mode $m$ and admissible
intervention set $\mathcal{R}_m \subseteq \mathcal{R}$.

\STATE \textbf{Evaluate interventions:} For each
$x_i \in \mathcal{D}_{\text{train}}$ and intervention
$r_j \in \mathcal{R}_m$, record success $s(x_i,r_j)$
and cost $c(x_i,r_j)$.

\STATE \textbf{Distill policy:} Enumerate bounded candidate policies
$\pi \in \Sigma(\mathcal{R}_m)$ and score them by $J(\pi)$.

\STATE \textbf{Freeze:} Deploy
$\pi^* = \arg\max_{\pi} J(\pi)$
as the task-family recovery policy.

\end{algorithmic}

\end{algorithm}

\textbf{Candidate intervention library.}
The candidate library $\mathcal{R}$ is fixed before test evaluation. It contains the recovery interventions available to all relevant baselines in a benchmark family: action guards for ALFWorld, procedural knowledge sources and local induction rules for AppWorld, and retrieval-demonstration budgets for Finance. DARC does not invent new interventions at test time; it restricts this pre-specified library to the interventions that match the diagnosed failure mode.

\subsection{Failure Diagnosis}
\label{sec:diagnosis}

DARC converts development-set failure analysis into a harness restriction. For each task family, we execute the base agent on development tasks and measure observable failure signatures: invalid or state-incompatible actions in ALFWorld, missing cross-application procedures in AppWorld, and exact-format violations in XBRL Finance. These signatures identify a dominant failure mode and determine the admissible intervention set $\mathcal{R}_m$.

This frozen, task-family-level diagnosis makes the harness auditable, avoids test-time updates, and prevents the policy from observing test labels. While the framework can support learned per-instance routing, our experiments evaluate the controlled setting where each family instantiates one dominant failure mode. Appendix~\ref{app:mapping} lists the failure signals and intervention sets, while Appendix~\ref{app:diagnosis_records} provides representative development-trace diagnosis examples.

The output of diagnosis is not a longer prompt; it is a constrained recovery space. ALFWorld receives an action-validity harness, AppWorld receives a procedural-knowledge harness, and Finance receives a format-precision harness based on retrieval demonstrations. This restriction is the key difference from a universal playbook: interventions that cannot address the diagnosed failure mode are excluded before policy search, reducing irrelevant or conflicting recovery context.

\subsection{Recovery-Policy Distillation}
\label{sec:policy_distillation}

Given $\mathcal{R}_m$, DARC estimates which intervention or intervention sequence should be used at deployment time. For each training task $x_i$ and intervention $r_j$, we collect
\begin{equation}
E_{i,j} = \big(s(x_i,r_j), c(x_i,r_j)\big),
\end{equation}
where success is measured by the same task verifier used for training-set adaptation and cost is measured in the relevant budget unit. This uses training-set success feedback; it does not use test labels. Section~\ref{sec:experiments} reports this accounting explicitly for all methods.

DARC evaluates a candidate policy $\pi=(r_1,\ldots,r_L)$ as a short-circuit policy: the agent tries interventions in order and stops when one succeeds. The halting index is
\begin{equation}
h_i^*(\pi)=
\begin{cases}
\min \{t: s(x_i,r_t)=1\}, & \text{if any intervention in } \pi \text{ succeeds,}\\
L, & \text{otherwise.}
\end{cases}
\end{equation}
The empirical success and cost are
\begin{equation}
\widehat{\mathrm{succ}}(\pi)=\frac{1}{|\mathcal{D}_{\text{train}}|}\sum_i \max_{r_t \in \pi}s(x_i,r_t),
\end{equation}
\begin{equation}
\widehat{\mathrm{cost}}(\pi)=\frac{1}{|\mathcal{D}_{\text{train}}|}\sum_i \sum_{t=1}^{h_i^*(\pi)}c(x_i,r_t).
\end{equation}
The deployed policy maximizes
\begin{equation}
\label{eq:darc_objective}
J(\pi)=\widehat{\mathrm{succ}}(\pi)-\lambda \max(0,\widehat{\mathrm{cost}}(\pi)-\tau_{\text{free}}).
\end{equation}

The short-circuit estimate assumes that intervention attempts are evaluated under a controlled retry protocol: later interventions are invoked only after an earlier intervention fails, and the verifier can attribute success and cost to the invoked intervention. In resettable environments, each intervention can be evaluated from the same initial state; in retrieval and extraction settings, interventions differ only in the supplied evidence budget. This assumption is explicit because it determines when an offline evidence matrix is a valid proxy for deployment. Table~\ref{tab:deployment-semantics} summarizes the deployment semantics used by each benchmark family.

\begin{table*}[t]
\centering
\caption{Deployment semantics for the reported DARC policies. The policy is frozen before test evaluation; later interventions are invoked only after earlier interventions fail under the benchmark verifier.}
\label{tab:deployment-semantics}
\begin{tabular}{lp{5.5cm}p{5cm}p{4cm}}
\toprule
\textbf{Benchmark} & \textbf{Later intervention observes} & \textbf{Side-effect control} & \textbf{Reported cost} \\
\midrule
\rowcolor{blue!5} ALFWorld & Same episode specification with a stricter action-validity harness & Resettable evaluation from the same initial environment state & Environment steps, invalid actions, seconds per episode \\
\rowcolor{green!5} AppWorld & Richer procedural context after a cheaper procedural source fails & Controlled task attempts with success attributed by the task verifier & Environment steps and task/scenario completion \\
\rowcolor{orange!5} Finance & Same extraction query with a different retrieval-demonstration budget & Stateless extraction; no environment side effects & Retrieval budget $k$ and answer accuracy \\
\bottomrule
\end{tabular}
\end{table*}

\subsection{Structural Properties}
\label{sec:theory}

The success term in DARC is a coverage objective over recovery interventions. For a subset $A \subseteq \mathcal{R}_m$, define $f(A)=\frac{1}{n}\sum_i \max_{r\in A}s(x_i,r)$. Then $f$ is monotone and submodular, because adding an intervention can only cover additional tasks and marginal gains shrink as the covered set grows. This explains why short policies can capture most of the reachable recovery set in our finite candidate spaces.

The full objective in Eq.~\ref{eq:darc_objective} is an ordered, cost-penalized policy objective, so the coverage result alone does not establish optimal cost-aware ordering. Instead, diagnosis reduces the candidate space and policy distillation enumerates bounded policies inside that space. For a fixed finite policy class $\Sigma_K$, standard uniform convergence gives that the empirical maximizer approaches the best population policy at rate $O(\sqrt{\log |\Sigma_K|/n})$ under bounded per-task costs. Since $|\Sigma_K|$ decreases with $|\mathcal{R}_m|$, diagnosis-guided restriction supports selection within a fixed finite candidate class; transfer across new failure modes remains an empirical question.

%% file: sections/04-experiments.tex
\section{Experiments}
\label{sec:experiments}

We evaluate whether diagnosis-guided recovery improves agent self-correction under controlled adaptation budgets. The experiments answer four questions:
\begin{itemize}[leftmargin=*,itemsep=0pt,parsep=0pt,topsep=4pt]
    \item \textbf{RQ1 (Performance):} How does DARC compare with base agents and generic playbook baselines?
    \item \textbf{RQ2 (Transfer):} Do frozen recovery policies transfer across related splits or tasks?
    \item \textbf{RQ3 (Diagnosis):} How much does selecting the correct diagnosis-guided recovery policy matter relative to generic or mismatched policies?
    \item \textbf{RQ4 (Cost):} Does the distilled policy reduce unnecessary environment steps or retrieval budget?
\end{itemize}

\subsection{Experimental Setup}

\textbf{Benchmarks and recovery diagnoses.}
We instantiate one dominant failure diagnosis per benchmark family (Appendix~\ref{app:mapping}). \textbf{ALFWorld}~\cite{shridhar2020alfworld} is treated as an action-validity setting, so the recovery harness contains task-specific action guards. \textbf{AppWorld}~\cite{trivedi2024appworld} is treated as a procedural-knowledge setting, so the harness contains Auto-Knowledge, local induction, and retrieval fallback interventions. \textbf{XBRL Finance}~\cite{loukas2022finer,wang2025finlora} is treated as a format-precision setting, so the harness restricts interventions to retrieval-based demonstrations and distills the retrieval budget $k$.

\textbf{Baselines.}
We compare against \textbf{Base LLM}, \textbf{ICL}~\cite{agarwal2024many}, \textbf{MIPROv2}~\cite{opsahl2024optimizing}, \textbf{GEPA}~\cite{agrawal2025gepa}, and \textbf{ACE}~\cite{zhang2026agentic}. These baselines cover many-shot in-context learning, instruction/demonstration optimization, and reflective prompt evolution~\cite{khattab2023dspy}. All offline-adaptation methods use the same training split and adaptation budget available to DARC unless a baseline is not applicable to a benchmark. Appendix~\ref{app:baselines} summarizes the mechanisms.

\textbf{Supervision and feedback accounting.}
DARC uses training-set verifier outcomes to estimate intervention success in the evidence matrix; it does not use test labels. This differs from prompt-optimization baselines that directly use answer labels to construct prompts or demonstrations. Table~\ref{tab:feedback-accounting} separates these forms of information.

\begin{table}[t]
\centering
\caption{Information used during adaptation. ``Train success'' denotes training-set verifier or task-success feedback used to score candidate policies. ``Answer labels'' denotes direct use of ground-truth answers to construct prompts, demonstrations, or optimized instructions. No method uses test labels.}
\label{tab:feedback-accounting}
\resizebox{\columnwidth}{!}{
\begin{tabular}{lccc}
\toprule
\textbf{Method} & \textbf{Train success} & \textbf{Answer labels} & \textbf{Test labels} \\
\midrule
Base LLM & No & No & No \\
ICL & No & Yes & No \\
MIPROv2 / GEPA & Yes & Yes & No \\
ACE & Yes & No & No \\
DARC & Yes & No & No \\
\bottomrule
\end{tabular}
}
\end{table}

\textbf{Evaluation scope.} We report all quantitative metrics as empirical estimates across the respective evaluation splits, and we complement the headline comparison with scenario-cluster bootstrap confidence intervals and paired significance tests to assess whether the observed gaps exceed split-level sampling noise (Figure~\ref{fig:appworld_significance}). To isolate the impact of our diagnosis-guided restriction, the ablation studies compare three recovery strategies: (1) the \textit{correct} policy, which aligns interventions with the diagnosed failure mode; (2) a \textit{generic} baseline that uses a standard, unconstrained recovery playbook; and (3) a \textit{mismatched} policy that acts as a negative control by applying an incompatible intervention set. This design separates the actual benefits of targeted recovery from the simple effect of adding more context.

\subsection{Main Benchmark Results}

Tables~\ref{tab:main-results} and~\ref{tab:finance-results} report the main results. Across the evaluated settings, DARC generally improves average performance over base agents and generic playbooks, but the gains are not uniform across every metric. In particular, ACE matches or exceeds DARC on some AppWorld challenge metrics for Qwen3.5-27B and Qwen3.6-27B~\cite{qwen3.5,qwen3.6-27b}. We therefore interpret these results as evidence for the average utility of diagnosis-guided recovery, though the gains vary by metric.

On ALFWorld, the action-validity recovery policy substantially improves both seen and unseen splits. With DeepSeek-V4-Flash, valid-unseen success increases from 39.55\% for the base agent and 54.48\% for ACE to 90.30\%. On AppWorld, the procedural recovery policy gives the strongest gains on Test-Normal and improves average TGC for all three base models, while challenge-split SGC remains competitive with ACE rather than uniformly better. On Finance, the format-precision policy reaches 94.50\% macro accuracy with DeepSeek-V4-Flash, compared with 80.50\% for ACE and 74.00\% for MIPROv2 under the reported protocol. As detailed in Table~\ref{tab:finance-results}, ``Answer Labels'' follows Table~\ref{tab:feedback-accounting}: whether ground-truth answers are directly used to construct prompts or demonstrations during adaptation. DARC uses training-set verifier outcomes to score retrieval budgets but does not use test labels.

\begin{figure}[t]
\centering
\includegraphics[width=0.95\columnwidth]{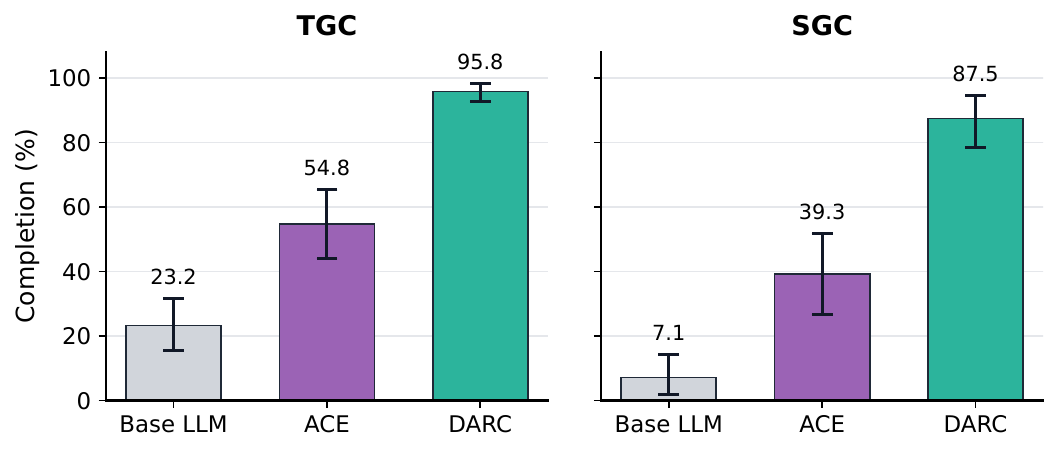}
\caption{AppWorld Test-Normal completion for DeepSeek-V4-Flash with 95\% scenario-cluster bootstrap confidence intervals (20{,}000 resamples over 56 scenarios). The intervals of DARC do not overlap those of ACE or the base agent on either TGC or SGC, and paired scenario-level tests with Holm correction confirm the gaps are significant.}
\label{fig:appworld_significance}
\end{figure}

\textbf{Statistical significance.}
Because agent benchmarks are evaluated on finite task splits, we quantify the uncertainty of the headline AppWorld comparison at the scenario level. AppWorld Test-Normal contains 168 tasks nested within 56 scenarios (three tasks per scenario); since outcomes within the same scenario are correlated, we treat the scenario rather than the individual task as the independent statistical unit. We align all methods by task identifier, compute per-scenario TGC as the fraction of completed tasks and SGC as an indicator that all three tasks are completed, and estimate 95\% confidence intervals with 20{,}000 scenario-cluster bootstrap resamples shared across methods. Figure~\ref{fig:appworld_significance} reports the resulting intervals for DeepSeek-V4-Flash: those of DARC do not overlap those of ACE or the base agent. Using a two-sided paired $t$-test for per-scenario TGC and an exact McNemar test for paired SGC outcomes, we confirm that all DARC-versus-baseline comparisons remain significant across the 56 scenarios after Holm correction. These intervals quantify scenario-sampling uncertainty rather than variation across model-training seeds.

\begin{table*}[t]
\centering
\caption{Benchmark results on ALFWorld and AppWorld. For ALFWorld, we report success rates on valid\_seen and valid\_unseen. For AppWorld, TGC is Task Goal Completion and SGC is Scenario Goal Completion. Bold marks the best value within each base-model block and metric.}
\label{tab:main-results}
\resizebox{0.8\textwidth}{!}{
\begin{tabular}{l|ccc|cc|cc|cc}
\toprule
\multirow{3}{*}{\textbf{Method}} & \multicolumn{3}{c|}{\textbf{ALFWorld}} & \multicolumn{6}{c}{\textbf{AppWorld}} \\
\cmidrule{2-10}
& \multirow{2}{*}{\textbf{valid\_seen$\uparrow$}} & \multirow{2}{*}{\textbf{valid\_unseen$\uparrow$}} & \multirow{2}{*}{\textbf{Macro Avg.$\uparrow$}} & \multicolumn{2}{c|}{\textbf{Test-Normal}} & \multicolumn{2}{c|}{\textbf{Test-Challenge}} & \multicolumn{2}{c}{\textbf{Average}} \\
\cmidrule{5-10}
& & & & \textbf{TGC$\uparrow$} & \textbf{SGC$\uparrow$} & \textbf{TGC$\uparrow$} & \textbf{SGC$\uparrow$} & \textbf{TGC$\uparrow$} & \textbf{SGC$\uparrow$} \\
\midrule
\rowcolor{gray!15} \multicolumn{10}{c}{DeepSeek-V4-Flash} \\
Base LLM & 40.71\% & 39.55\% & 40.13\% & 23.21\% & 7.14\% & 13.43\% & 6.47\% & 18.32\% & 6.81\% \\
ICL & 46.43\% & 42.54\% & 44.48\% & 59.52\% & 42.86\% & 37.17\% & 19.42\% & 48.35\% & 31.14\% \\
MIPROv2 & 54.29\% & 49.25\% & 51.77\% & 57.14\% & 35.71\% & 32.85\% & 10.07\% & 45.00\% & 22.89\% \\
GEPA & 59.29\% & 63.43\% & 61.36\% & 54.17\% & 30.36\% & 33.57\% & 12.95\% & 43.87\% & 21.65\% \\
ACE & 50.00\% & 54.48\% & 52.24\% & 54.76\% & 39.29\% & 35.73\% & 17.99\% & 45.25\% & 28.64\% \\
DARC & \textbf{93.57\%} & \textbf{90.30\%} & \textbf{91.94\%} & \textbf{95.83\%} & \textbf{87.50\%} & \textbf{53.96\%} & \textbf{30.22\%} & \textbf{74.90\%} & \textbf{58.86\%} \\
\midrule
\rowcolor{gray!15} \multicolumn{10}{c}{Qwen3.5-27B} \\
Base LLM & 58.57\% & 55.22\% & 56.90\% & 42.26\% & 23.21\% & 23.74\% & 12.23\% & 33.00\% & 17.72\% \\
ICL & 72.14\% & 70.90\% & 71.52\% & 75.00\% & 57.14\% & 59.71\% & 38.13\% & 67.36\% & 47.64\% \\
MIPROv2 & 69.29\% & 60.45\% & 64.87\% & 4.76\% & 0.00\% & 6.95\% & 0.72\% & 5.86\% & 0.36\% \\
GEPA & 79.29\% & 72.39\% & 75.84\% & 50.60\% & 17.86\% & 34.29\% & 12.95\% & 42.44\% & 15.40\% \\
ACE & 80.00\% & 81.34\% & 80.67\% & 79.17\% & 64.29\% & 74.58\% & \textbf{51.08\%} & 76.88\% & 57.68\% \\
DARC & \textbf{94.29\%} & \textbf{95.52\%} & \textbf{94.90\%} & \textbf{93.45\%} & \textbf{89.29\%} & \textbf{65.47\%} & 41.01\% & \textbf{79.46\%} & \textbf{65.15\%} \\
\midrule
\rowcolor{gray!15} \multicolumn{10}{c}{Qwen3.6-27B} \\
Base LLM & 59.29\% & 56.72\% & 58.00\% & 42.86\% & 32.14\% & 26.14\% & 12.95\% & 34.50\% & 22.55\% \\
ICL & 75.00\% & 76.12\% & 75.56\% & 60.71\% & 46.43\% & 41.73\% & 25.18\% & 51.22\% & 35.80\% \\
MIPROv2 & 63.57\% & 55.22\% & 59.40\% & 14.29\% & 8.93\% & 6.47\% & 2.88\% & 10.38\% & 5.91\% \\
GEPA & 90.00\% & 83.58\% & 86.79\% & 45.83\% & 30.36\% & 34.53\% & 18.71\% & 40.18\% & 24.53\% \\
ACE & 80.71\% & 78.36\% & 79.54\% & 67.86\% & 55.36\% & \textbf{53.00\%} & \textbf{32.37\%} & 60.43\% & 43.86\% \\
DARC & \textbf{97.14\%} & \textbf{95.52\%} & \textbf{96.33\%} & \textbf{86.31\%} & \textbf{78.57\%} & 47.72\% & \textbf{32.37\%} & \textbf{67.02\%} & \textbf{55.47\%} \\
\bottomrule
\end{tabular}
}
\end{table*}

\begin{table}[t]
\centering
\caption{Finance benchmark results on FiNER tags and Formula extraction.}
\label{tab:finance-results}
\resizebox{\columnwidth}{!}{
\begin{tabular}{lc|ccc}
\toprule
\textbf{Method} & \textbf{Answer Labels} & \textbf{FiNER / Tags Acc$\uparrow$} & \textbf{Formula Acc$\uparrow$} & \textbf{Average$\uparrow$} \\
\midrule
\rowcolor{gray!15} \multicolumn{5}{c}{DeepSeek-V4-Flash} \\
Base LLM & No & 68.00\% & 8.00\% & 38.00\% \\
ICL & Yes & 77.50\% & 66.00\% & 71.75\% \\
MIPROv2 & Yes & 78.50\% & 69.50\% & 74.00\% \\
GEPA & Yes & 82.50\% & 61.00\% & 71.75\% \\
ACE & No & 83.00\% & 78.00\% & 80.50\% \\
DARC & No & \textbf{90.00\%} & \textbf{99.00\%} & \textbf{94.50\%} \\
\midrule
\rowcolor{gray!15} \multicolumn{5}{c}{Qwen3.5-27B} \\
Base LLM & No & 47.50\% & 24.00\% & 35.75\% \\
ICL & Yes & 69.50\% & 45.50\% & 57.50\% \\
MIPROv2 & Yes & 72.00\% & 58.00\% & 65.00\% \\
GEPA & Yes & 68.00\% & 61.00\% & 64.50\% \\
ACE & No & 82.00\% & 79.00\% & 80.50\% \\
DARC & No & \textbf{87.50\%} & \textbf{98.50\%} & \textbf{93.00\%} \\
\midrule
\rowcolor{gray!15} \multicolumn{5}{c}{Qwen3.6-27B} \\
Base LLM & No & 53.00\% & 36.00\% & 44.50\% \\
ICL & Yes & 67.00\% & 46.00\% & 56.50\% \\
MIPROv2 & Yes & 72.50\% & 60.00\% & 66.25\% \\
GEPA & Yes & 77.50\% & 60.00\% & 68.75\% \\
ACE & No & 81.50\% & 75.00\% & 78.25\% \\
DARC & No & \textbf{86.50\%} & \textbf{98.50\%} & \textbf{92.50\%} \\
\bottomrule
\end{tabular}
}
\end{table}

\subsection{Cross-Task Transferability}

We next test whether a frozen recovery policy can transfer across related tasks. For Finance, we distill a retrieval-budget policy on Formula and deploy it on FiNER tags, and vice versa. For ALFWorld, we distill on valid\_seen and deploy on valid\_unseen. As shown in Table~\ref{tab:cross-task}, the target-tuned row is shown only when a target-specific policy was available.

\begin{table}[h]
  \centering
  \caption{Cross-task transferability of frozen DARC policies. Finance transfers between the Formula and FiNER-tags tasks; ALFWorld transfers from valid\_seen to valid\_unseen.}
  \label{tab:cross-task}
  \resizebox{\columnwidth}{!}{
  \begin{tabular}{lccc}
  \toprule
  & \multicolumn{2}{c}{\textbf{Finance (Acc$\uparrow$)}} & \textbf{ALFWorld (Succ$\uparrow$)} \\
  \cmidrule(lr){2-3} \cmidrule(lr){4-4}
  \textbf{Method} & Formula $\rightarrow$ FiNER & FiNER $\rightarrow$ Formula & seen $\rightarrow$ unseen \\
  \midrule
  Base LLM & 68.00\% & 8.00\% & 39.55\% \\
  ICL & 77.50\% & 66.00\% & 42.54\% \\
  MIPROv2 & 78.50\% & 69.50\% & 49.25\% \\
  GEPA & 82.50\% & 61.00\% & 63.43\% \\
  ACE & 83.00\% & 78.00\% & 54.48\% \\
  \midrule
  DARC (Frozen) & \textbf{89.50\%} & \textbf{96.00\%} & \textbf{90.30\%} \\
  DARC (Target-Tuned) & 90.00\% & 99.00\% & -- \\
  \bottomrule
  \end{tabular}
  }
\end{table}

These within-family transfer results suggest that the frozen recovery policy can preserve much of the target-tuned performance when the failure mode remains similar. For Finance, the frozen retrieval-budget policy remains close to the target-tuned policy: Formula-to-FiNER reaches 89.50\% versus 90.00\%, and FiNER-to-Formula reaches 96.00\% versus 99.00\%. For ALFWorld, the same action-validity policy transfers from valid\_seen to valid\_unseen and reaches 90.30\%.

\section{Analysis and Ablation}
\label{sec:analysis}

The diagnosis-guided view predicts that recovery interventions are not interchangeable across failure modes. A retrieval demonstration can help Finance formatting but cannot make an ALFWorld action admissible. An action guard can prune embodied commands but cannot supply the AppWorld API procedure needed for a multi-application workflow. We test this prediction by comparing correct, generic, and mismatched recovery policies.

\subsection{Ablation: Recovery Diagnosis}

\begin{table}[h]
\centering
\caption{Impact of selecting the correct recovery diagnosis.}
\label{tab:routing-ablation}
\resizebox{\columnwidth}{!}{
\begin{tabular}{llccc}
\toprule
\textbf{Benchmark} & \textbf{Metric} & \textbf{Correct} & \textbf{Generic} & \textbf{Mismatched} \\
\midrule
ALFWorld & Success$\uparrow$ & 91.94\% & 52.24\% & 52.24\% \\
AppWorld & TGC$\uparrow$ & 70.24\% & 61.90\% & 64.88\% \\
AppWorld & SGC$\uparrow$ & 53.57\% & 50.00\% & 46.43\% \\
Finance & Macro$\uparrow$ & 94.50\% & 80.50\% & 37.75\% \\
\bottomrule
\end{tabular}
}
\end{table}

Table~\ref{tab:routing-ablation} shows that the correct recovery policy (aligned with the diagnosed failure mode) outperforms both a generic playbook and a mismatched intervention set across all evaluated settings. For AppWorld Test-Normal (evaluated under a stricter 25-step budget not comparable to Table~\ref{tab:main-results}), correct diagnosis reaches 70.24\% TGC compared to 61.90\% (generic) and 64.88\% (mismatched), while achieving the best SGC with fewer environment steps than the generic playbook (36.60 vs.\ 42.97).

While AppWorld confidence intervals overlap, the directional advantage is consistent. Notably, the mismatched policy exceeds the generic playbook on AppWorld TGC. This supports our premise that excessive irrelevant context (generic) can be more harmful than largely inert mismatched interventions, though the expected ordering holds under the stricter SGC metric. The penalty for mismatched interventions is much larger on ALFWorld and Finance, where they actively conflict with the required recovery behavior.

Overall, these results highlight the necessity of diagnosis-guided restriction: adding context only helps when interventions match the underlying failure mode. Further ablations (Appendix~\ref{app:diagnosis-isolation}) confirm that isolating the diagnosis step maintains accuracy while significantly improving search efficiency and stability.

\begin{figure}[h]
\centering
\includegraphics[width=0.95\columnwidth]{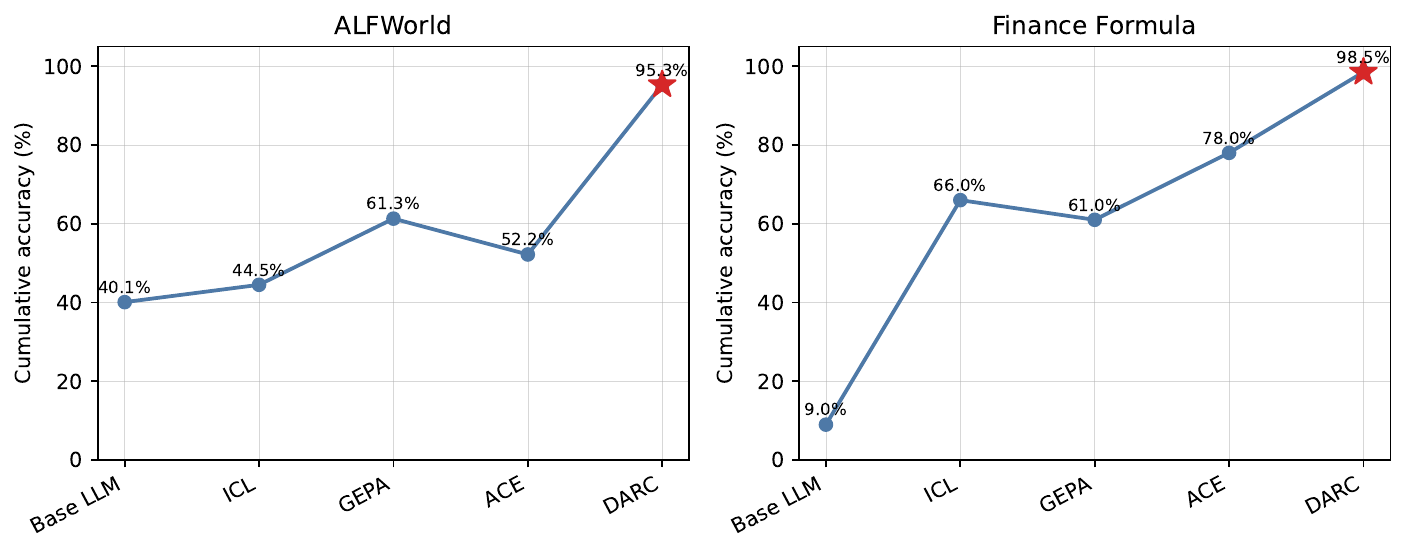}
\caption{Cumulative solved episodes on ALFWorld as the maximum environment-step budget increases. DARC solves more episodes at lower budgets because the action-validity recovery policy reduces invalid or inefficient actions.}
\label{fig:cumulative_perf}
\end{figure}

\subsection{Matched-Information Fairness}
\label{sec:fairness}

The large ALFWorld gain invites a fairness concern: does DARC win only because its action-validity guard sees privileged admissibility information that baselines do not? We test this directly on the full 134-task valid\_unseen split, with all methods using the same DeepSeek-V4-Flash backbone, identical environment observations, the full set of admissible actions returned by the environment, a 50-step budget, and the same evaluation script. Crucially, DARC's guard is not a model-probability ranker and does not read the environment's ground-truth action: it is a deterministic, diagnosis-aware rule scorer that ranks each admissible command $a$ by
\begin{equation}
\label{eq:guard_score}
S(a)=S_{\text{goal}}+S_{\text{phase}}+S_{\text{search}}+S_{\text{train-prior}}-S_{\text{conflict}}-S_{\text{loop}},
\end{equation}
and keeps the top 12. All terms are computed from the task description and training-set priors; none use test labels or environment-provided correct actions.

Table~\ref{tab:matched-fairness} decomposes the sources of the gain. In this table, ``Base + matched metadata'' adds only parsed task type, target object, and movable/goal containers on top of the full action set, without diagnosis, ranking, or direct execution. Numbers are a same-day matched rerun under the 50-step budget; the ranked-recovery row is the run reported as DARC valid\_unseen in Table~\ref{tab:main-results} (90.30\%). They are not directly comparable to the valid\_seen-selected frozen cascade in Table~\ref{tab:diagnosis-isolation} (99.25\%, different selection protocol). Granting the base agent the same parsed metadata (task type, target object, movable and goal containers) improves success only marginally (39.55\%$\rightarrow$43.28\%), so information access alone is not the driver. Restricting the action view to a \emph{random} top-12 collapses ACE (54.48\%$\rightarrow$25.37\%), so action-space restriction by itself is not a free lunch and can hurt. Diagnosis as a bare label, without the ranked recovery operator, stays at base level (38.81\%). The gain materializes only with the ranked recovery operator, which reaches 89--90\% while cutting invalid actions per episode to 0.11. The ALFWorld improvement is therefore attributable to the quality of the diagnosis-guided \emph{ranking} of already-available admissible actions, not to privileged information or restriction alone.

Appendix~\ref{app:factorial} isolates this further with a $2\times2$ factorial over the two components of the harness. Neither component works alone: the recovery prompt on the full action set is worth $-0.75$ pp, the ranked action view without the recovery prompt is worth $+3.73$ pp, and forcing the guard's top-1 command with no language model in the loop reaches only 40.33\%. The two components together are worth $+49.26$ pp, an interaction of $+46.27$ pp. The ALFWorld gain is thus a property of the recovery \emph{interface}, where restriction and instruction are explicitly matched to each other, rather than of either prompt content or action pruning in isolation.

\begin{table}[h]
\centering
\caption{Matched-information fairness evaluation on the ALFWorld valid\_unseen split.}
\label{tab:matched-fairness}
\resizebox{\columnwidth}{!}{
\begin{tabular}{llrrrr}
\toprule
\textbf{Method} & \textbf{Action view} & \textbf{Success$\uparrow$} & \textbf{Steps$\downarrow$} & \textbf{Invalid/ep$\downarrow$} & \textbf{vs.\ ACE} \\
\midrule
Base LLM & Full & 39.55\% & 35.25 & 1.709 & $-14.93$ \\
Base + matched metadata & Full & 43.28\% & 34.02 & 1.560 & $-11.19$ \\
ACE & Full & 54.48\% & 30.06 & 1.784 & $0.00$ \\
ACE + random top-$k$ & Random top-12 & 25.37\% & 42.16 & 7.403 & $-29.10$ \\
DARC, diagnosis only & Full & 38.81\% & 34.69 & 2.507 & $-15.67$ \\
DARC, ranked recovery & Ranked top-12 & \textbf{90.30\%} & \textbf{16.50} & 0.575 & $+35.82$ \\
DARC, full guard & Ranked top-12 & 89.55\% & 18.38 & \textbf{0.112} & $+35.07$ \\
\bottomrule
\end{tabular}
}
\end{table}

\subsection{Weight-Space Training Extension}

Beyond test-time recovery, we conduct a preliminary probe into extending DARC to weight-space training. Using Qwen3-8B, we compare two full-parameter fine-tuning methods guided by a DARC-derived two-stage curriculum (\texttt{diverse\_replay} then \texttt{uniform\_rollout}, allocating 5.0M and 2.5M tokens respectively). \textbf{DARC-GRPO} updates weights using standard GRPO, while \textbf{DARC-OPSD} adds an online exponential-moving-average (EMA) teacher for policy self-distillation. Table~\ref{tab:opsd-training-extension} summarizes the results for the final checkpoints.

\begin{table}[h]
\centering
\caption{Full evaluation of weight-space training extensions on ALFWorld. Models are fine-tuned from Qwen3-8B using a DARC-derived two-stage curriculum.}
\label{tab:opsd-training-extension}
\resizebox{\columnwidth}{!}{
\begin{tabular}{lrrrr}
\toprule
\textbf{Method} & \textbf{valid\_seen} & \textbf{valid\_unseen} & \textbf{Macro$\uparrow$} & \textbf{$\Delta$ Base} \\
\midrule
Base Qwen3-8B & 7.14\% & 4.48\% & 5.81\% & -- \\
DARC-GRPO & \textbf{12.14\%} & \textbf{8.96\%} & \textbf{10.55\%} & \textbf{+4.74 pp} \\
DARC-OPSD & 9.29\% & 6.72\% & 8.00\% & +2.19 pp \\
\bottomrule
\end{tabular}
}
\end{table}

Both weight-space variants improve success over the base model, suggesting that the DARC-derived curriculum transfers to weight-space training. DARC-GRPO improves macro success by 4.74 percentage points, while DARC-OPSD yields a smaller 2.19 pp gain, indicating that online self-distillation does not provide additional benefit over standard GRPO in this setting. Because these are single-seed runs with modest absolute gains, we treat this as a proof of concept rather than a fully verified training recipe.

\subsection{Benchmark-Specific Failure Modes}

\textbf{Action validity in ALFWorld.}
The ALFWorld recovery policy restricts the agent to state-compatible actions. With DeepSeek-V4-Flash, DARC reduces invalid actions per episode to 0.091 (Table~\ref{tab:cost-speed}) and improves macro success to 91.94\% (Table~\ref{tab:main-results}). This supports the interpretation that action validity is the dominant failure mode in this setting.

\textbf{Procedure breadth in AppWorld.}
AppWorld requires multi-step API knowledge. The procedural recovery policy uses a fallback chain because the interventions are complementary: a local induction rule can recover tasks that an Auto-Knowledge source misses. The gains are strongest on Test-Normal; the Test-Challenge split remains harder and includes metrics where ACE is competitive or better.

\textbf{Format precision in Finance.}
Finance shows that a recovery policy may tune a single intervention dimension rather than switch among heterogeneous interventions. Once the admissible set is restricted to retrieval demonstrations, DARC selects how much evidence to retrieve. Table~\ref{tab:finance-policy} shows that larger retrieval budgets are not monotonic: $k=8$ and $k=16$ slightly reduce tag accuracy relative to smaller budgets. The distilled policy uses $k=2$ for Formula and $k=1$ for Tags, staying in the top macro-accuracy band while reducing the mean retrieval budget to 1.5 demonstrations.

\begin{table}[h]
\centering
\caption{Finance retrieval-budget distillation. The DARC policy selects $k=2$ for Formula and $k=1$ for Tags, staying in the top macro-accuracy band with a mean budget of 1.5 demonstrations.}
\label{tab:finance-policy}
\resizebox{\columnwidth}{!}{
\begin{tabular}{llrrrr}
\toprule
\textbf{Method} & \textbf{Answer Labels} & \textbf{Formula} & \textbf{Tags} & \textbf{Macro} & \textbf{Mean $k$} \\
\midrule
Base LLM & No & 8.0\% & 68.0\% & 38.00\% & -- \\
ICL & Yes & 66.0\% & 77.5\% & 71.75\% & 4.0 \\
MIPROv2 & Yes & 69.5\% & 78.5\% & 74.00\% & -- \\
GEPA & Yes & 61.0\% & 82.5\% & 71.75\% & -- \\
ACE & No & 78.0\% & 83.0\% & 80.50\% & -- \\
\midrule
Fixed retrieval $k=1$ & No & 96.0\% & 90.0\% & 93.00\% & 1.0 \\
Fixed retrieval $k=2$ & No & 99.0\% & 89.5\% & 94.25\% & 2.0 \\
Fixed retrieval $k=4$ & No & 99.0\% & 89.5\% & 94.25\% & 4.0 \\
Fixed retrieval $k=8$ & No & 99.0\% & 89.0\% & 94.00\% & 8.0 \\
Fixed retrieval $k=16$ & No & 99.0\% & 88.0\% & 93.50\% & 16.0 \\
\midrule
DARC policy & No & \textbf{99.0\%} & \textbf{90.0\%} & \textbf{94.50\%} & \textbf{1.5} \\
\bottomrule
\end{tabular}
}
\end{table}

\subsection{Cost-Success Tradeoff}

\begin{figure}[h]
\centering
\includegraphics[width=0.88\columnwidth]{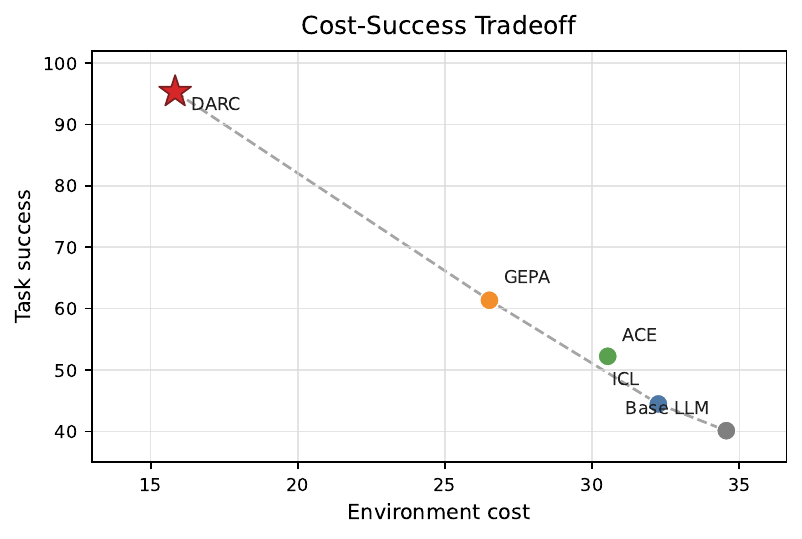}
\caption{Cost-success tradeoff on AppWorld. The red star marks the distilled DARC policy; points farther right use more environment steps.}
\label{fig:pareto}
\end{figure}

Figure~\ref{fig:pareto} visualizes the candidate AppWorld policies. A policy that stacks many interventions can improve coverage but increases mean environment steps. A policy that is too short saves cost but misses recoverable tasks. The selected DARC policy lies on the observed high-success, lower-cost frontier for this candidate set.

\subsection{Cost and Speed Efficiency}

\begin{table}[!h]
\centering
\caption{Cost and speed efficiency on ALFWorld with DeepSeek-V4-Flash. DARC improves success per 100 environment steps by reducing invalid actions and shortening trajectories.}
\label{tab:cost-speed}
\resizebox{\columnwidth}{!}{
\begin{tabular}{lrrrr}
\toprule
\textbf{Method} & \textbf{Env Steps$\downarrow$} & \textbf{Invalid/ep.$\downarrow$} & \textbf{Sec./ep.$\downarrow$} & \textbf{Succ./100 steps$\uparrow$} \\
\midrule
Base LLM & 34.56 & 1.288 & 155.4 & 1.16 \\
ICL & 32.25 & 2.453 & 23.3 & 1.38 \\
GEPA & 26.51 & 1.489 & 20.0 & 2.31 \\
ACE & 30.53 & 1.343 & 143.8 & 1.71 \\
\midrule
DARC & \textbf{15.83} & \textbf{0.091} & \textbf{10.3} & \textbf{6.02} \\
\bottomrule
\end{tabular}
}
\end{table}

Table~\ref{tab:cost-speed} shows the cost side of the ALFWorld recovery policy. Relative to the base agent, DARC reduces average environment steps by 54.2\% and invalid actions by 92.9\%. It also improves success per 100 steps from 1.16 to 6.02. Figure~\ref{fig:cumulative_perf} shows the budget-level view of the same effect: DARC solves more episodes than each baseline at every maximum environment-step budget. These numbers support the claim that diagnosis-guided recovery can improve both success and cost in an action-validity setting.

%% file: sections/05-conclusion.tex
\section{Conclusion and Limitations}

In this work, we introduced DARC, a framework that recasts agent self-correction from indiscriminate context expansion to diagnosis-guided recovery. By profiling task-family failure modes to restrict admissible interventions and distilling a cost-aware policy, DARC improves success across ALFWorld, AppWorld, and XBRL Finance while mitigating the overhead and interference of generic playbooks. Future work includes extending our static diagnosis to dynamic, instance-level routing for mixed-mode failures, developing more sample-efficient policy distillation methods, and expanding full-library cascade comparisons to broader domains.

%% file: sections/appendix.tex
\section{Theoretical Proofs}
\label{app:proofs}

\subsection{Proof of Monotone Submodularity}
For an intervention subset $A$, a task is covered if at least one intervention in $A$ succeeds on it. Thus $f(A)=\frac{1}{n}\sum_i \mathbf{1}[\exists r\in A: s(x_i,r)=1]$ is a standard coverage function. Coverage functions are monotone because adding an intervention cannot remove a covered task, and submodular because the marginal gain of adding an intervention decreases as the covered set grows.

\subsection{Uniform-Convergence Bound for a Fixed Chain Class}
Let $\Sigma_K$ be a finite class of duplicate-free chains of length at most $K$, and suppose per-task cost is bounded by $C_{\max}$. For a fixed chain $S$, Hoeffding's inequality bounds the deviation of both empirical success and empirical cost. Applying a union bound over $\Sigma_K$ yields a uniform deviation term of order $O(\sqrt{\log|\Sigma_K|/n})$ for the penalized objective. The empirical maximizer is therefore near the best chain in the fixed class up to twice this deviation. This bound does not replace empirical validation; it only explains why reducing $|\Sigma_K|$ can reduce finite-sample search risk.

\section{Extended Related Work}
\label{app:extended_related}

\textbf{Cascades and adaptive computation.}
DARC is structurally related to LLM cascades and adaptive computation. FrugalGPT~\cite{chen2023frugalgpt} and AutoMix~\cite{aggarwal2023automix} route among language models to reduce cost, while Adaptive-RAG~\cite{jeong2024adaptive} adjusts retrieval complexity. DARC instead routes among recovery interventions inside an agent harness. The base model is fixed; what changes is whether the agent receives an action guard, procedural source, retrieval evidence, or another admissible recovery operation.

\textbf{Broader evaluation context.}
The benchmarks in this paper sit within a broader evaluation ecosystem for language-model capability, instruction following, tool use, and interactive autonomy~\cite{hendrycks2021mmlu,srivastava2023bigbench,suzgun2023bbh,liang2023helm,liu2024agentbench,yao2024tau}. This ecosystem increasingly shifts from single-turn answer quality toward process observability: intermediate reasoning, external tool calls, state changes, and verifiable task completion~\cite{khot2023decomposed,park2023generativeagents,liu2024agentlite,fourney2024magentic}. DARC follows this shift but focuses on what to do after failure: it treats failures as signals for selecting recovery interventions rather than as final outcomes alone.

\section{Recovery-Harness Mapping}
\label{app:mapping}

Table~\ref{tab:recovery_mapping} summarizes the evaluated task-family mappings. The mapping is frozen before test evaluation and should be interpreted as the recovery-harness instantiation used in this paper, not as a learned universal diagnostic router.

\begin{table*}[h]
\centering
\caption{Frozen task-family to recovery-harness mapping. The diagnostic signal is measured on development tasks; the admissible intervention set is then fixed for test-time policy distillation and deployment.}
\label{tab:recovery_mapping}

\resizebox{\textwidth}{!}{
\begin{tabular}{llll}
\toprule
\textbf{Task Family} & \textbf{Development-Set Signal} & \textbf{Failure Mode} & \textbf{Admissible Intervention Set} \\
\midrule
ALFWorld & Invalid / state-incompatible actions & Action validity & Action pruning and admissibility guards \\
AppWorld & Missing API workflow / multi-step procedure failures & Procedure breadth & Auto-Knowledge, local induction, retrieval fallback \\
Finance & Exact formula/tag format errors & Format precision & Retrieval few-shot demonstrations and budget selection \\
\bottomrule
\end{tabular}
}

\end{table*}

\section{Representative Diagnosis Records}
\label{app:diagnosis_records}

Table~\ref{tab:diagnosis-examples} makes the diagnosis interface concrete. The diagnostic LLM receives only development-set trajectory summaries and verifier-visible failure signals, then maps the dominant recurring pattern to one label from the fixed failure-mode vocabulary. These task-family diagnoses and their associated intervention sets are frozen before test evaluation.

\begin{table*}[h]
\centering
\caption{Representative development-trace diagnosis examples. The trace summaries compress recurring observable failures; no test labels or hidden benchmark state are supplied to the diagnostic LLM.}
\label{tab:diagnosis-examples}
\begin{tabular}{
    >{\raggedright\arraybackslash}p{0.13\textwidth}
    >{\raggedright\arraybackslash}p{0.61\textwidth}
    >{\raggedright\arraybackslash}p{0.16\textwidth}}
\toprule
\textbf{Benchmark} &
\textbf{Observed Failure Trace} &
\textbf{LLM Diagnosis} \\
\midrule
ALFWorld &
The agent repeatedly issues actions that are unavailable in the current environment state, making no state progress across consecutive steps. &
Action validity \\

AppWorld &
API calls omit a required upstream dependency, such as resolving an entity or resource identifier before invoking the downstream cross-application operation. &
Procedure breadth \\

Finance &
The retrieved evidence supports the correct formula or tag, but the response contains prose, extra fields, or a malformed schema that fails exact-match evaluation. &
Format precision \\
\bottomrule
\end{tabular}
\end{table*}

\section{Generated Recovery Trace Example}
\label{app:trace}

\begin{figure}[t]
\centering
\includegraphics[width=1.0\columnwidth]{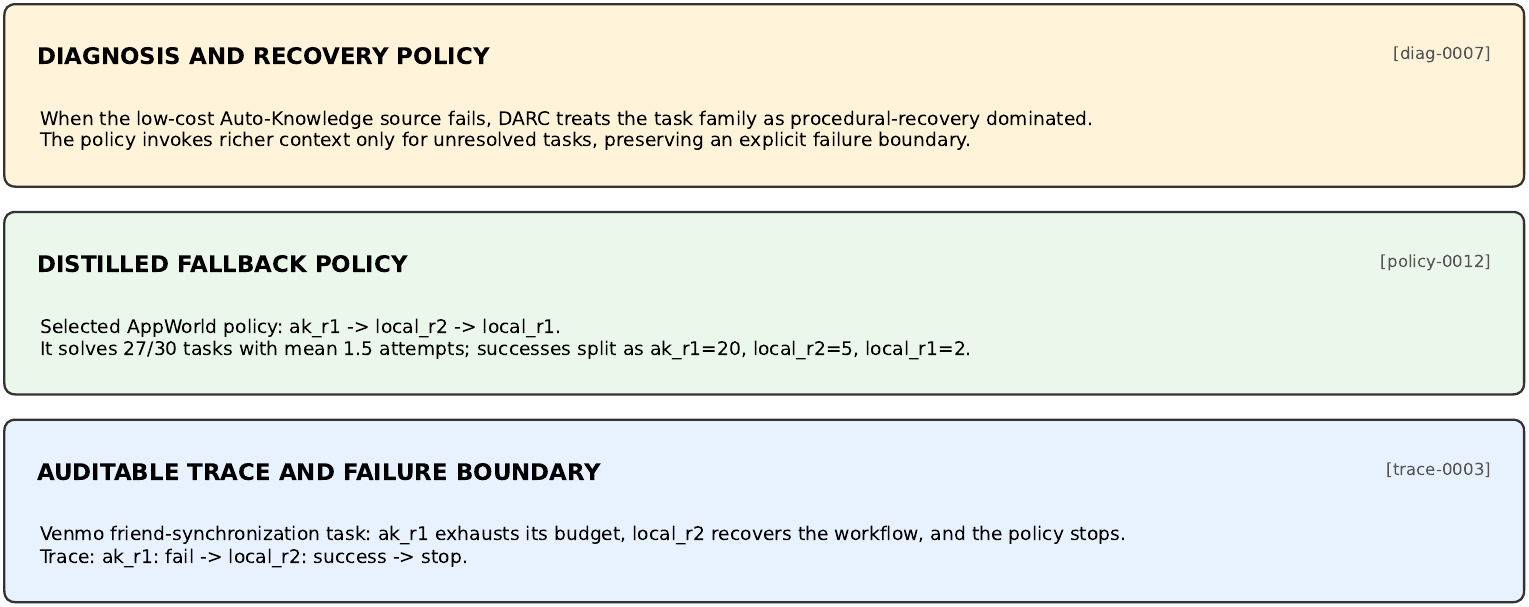}
\caption{Example AppWorld recovery trace. The distilled procedural policy invokes additional context only after a cheaper intervention fails.}
\label{fig:trace}
\end{figure}

Figure~\ref{fig:trace} illustrates the AppWorld procedural recovery harness. The distilled policy first uses a lower-cost procedural source and invokes richer context only when the earlier intervention fails, reducing unnecessary context calls while preserving recovery capacity.

\section{Factorial Decomposition of the ALFWorld Recovery Harness}
\label{app:factorial}

Section~\ref{sec:fairness} shows that neither privileged information nor action-space restriction alone explains the ALFWorld gain. This appendix reports the underlying $2\times2$ factorial experiment that separates the two components of the action-validity harness: the \emph{restricted action view} produced by the guard of Eq.~\ref{eq:guard_score}, and the \emph{recovery prompt} that instructs the agent how to act inside that view.

\textbf{Protocol.} All configurations run on the full ALFWorld valid\_unseen split (134 tasks) with DeepSeek-V4-Flash, the full set of admissible actions returned by the environment, a 50-step budget, temperature 0, and the same evaluation script. The two factors are toggled independently: the action view is either the full admissible set or the guard-ranked top-12, and the recovery prompt is either absent or present. We report exact task counts so that paired tests can be reproduced from the released per-task outcomes. Success is reported as solved task count over 134. Neither factor is effective in isolation; the gain appears only when the recovery prompt is paired with the restricted action view.

\begin{table}[h]
\centering
\caption{Factorial decomposition of the action-validity harness on ALFWorld valid\_unseen.}
\label{tab:factorial}
\resizebox{\columnwidth}{!}{
\begin{tabular}{llcc}
\toprule
\textbf{Action view} & \textbf{Recovery prompt} & \textbf{Solved} & \textbf{Success$\uparrow$} \\
\midrule
Full admissible set & Absent & 53 / 134 & 39.55\% \\
Full admissible set & Present & 52 / 134 & 38.81\% \\
Guard-ranked top-12 & Absent & 58 / 134 & 43.28\% \\
\rowcolor{blue!5} Guard-ranked top-12 & Present & \textbf{119 / 134} & \textbf{88.81\%} \\
\bottomrule
\end{tabular}
}
\end{table}

\textbf{The effect is an interaction, not a sum of main effects.}
Holding the action view at the full admissible set, adding the recovery prompt changes success by $-0.75$ pp ($53\rightarrow52$ tasks). Holding the recovery prompt absent, restricting the action view to the guard-ranked top-12 changes success by $+3.73$ pp ($53\rightarrow58$ tasks). The two single-factor effects therefore account for $+2.98$ pp in total, whereas enabling both factors yields $+49.26$ pp ($53\rightarrow119$ tasks). The difference-in-differences interaction term is
\begin{equation}
88.81 - 43.28 - 38.81 + 39.55 = +46.27~\text{pp},
\end{equation}
so essentially the entire effect is attributable to the pairing rather than to either component. This is the empirical form of the Intervention Mismatch argument in Section~\ref{sec:intro}: a recovery signal is only useful when it lands on an interface that can execute it.

\textbf{Invalid actions explain the mechanism.}
The three configurations for which we measured action-level statistics show why the prompt is inert without the restricted view. On the full admissible set, the recovery prompt \emph{increases} invalid actions per episode from 1.709 to 2.507 (Table~\ref{tab:matched-fairness}): the prompt tells the agent what the task requires, but many of the resulting commands are not executable in the current state, so the additional guidance is spent on rejected actions. Under the guard-ranked view the same guidance becomes executable and invalid actions fall to 0.575. The recovery prompt does not add knowledge the agent lacked; it becomes actionable only once the action interface is compatible with it.

\textbf{Auxiliary controls.}
Two further controls bound the contribution of the guard itself. First, forcing the top-1 guard-ranked command at every step \emph{without any language model} in the loop reaches 40.33\%, statistically indistinguishable from the 39.55\% base agent. The guard is therefore a high-recall candidate filter rather than a hand-written solver: its top-1 ordering is usually not the correct action, but the correct action is retained within the top-12 while distractors are removed. Second, attaching the same ranked view and recovery prompt to a plain base agent, without the remaining harness components, reaches 82.28\%, so most of the effect is carried by the restriction-prompt pairing and $6.53$ pp remain attributable to the rest of the harness.

\textbf{Scope of this decomposition.}
These are single-run results at temperature 0 on one split and one backbone, so the small single-factor effects ($-0.75$ pp and $+3.73$ pp, i.e.\ one and five tasks) should be read as null rather than as measured directions. The joint effect (66 tasks) is far outside this range. We report the decomposition for ALFWorld only; the corresponding factorial for the AppWorld procedural harness and the Finance retrieval harness is not available, and whether the same superadditive structure holds for those failure modes remains an open question.

\section{Baseline Descriptions}
\label{app:baselines}

We compare against two groups of methods:
\begin{itemize}
    \item \textbf{Base LLM}: The unaugmented model using a standard ReAct-style execution loop without additional recovery-harness construction.
    \item \textbf{ICL}: Standard prompting with a fixed set of demonstrations sampled or retrieved from the training set.
    \item \textbf{MIPROv2}: A DSPy prompt-optimization method that jointly tunes instructions and demonstrations.
    \item \textbf{GEPA}: A reflective prompt-evolution method that mutates prompts based on execution feedback.
    \item \textbf{ACE}: a broad recovery-playbook baseline constructed from training-set execution feedback.
\end{itemize}
All applicable offline-adaptation baselines use the same training split and adaptation budget as DARC. We further include a stronger matched baseline that performs validation-selected intervention-chain search without diagnosis-guided restriction (the full-library cascade in Table~\ref{tab:diagnosis-isolation}); extending it from ALFWorld to the other task families remains future work.

\section{Isolating the Diagnosis Step}
\label{app:diagnosis-isolation}

The generic and mismatched conditions in Section~\ref{sec:analysis} vary the intervention set but do not run the same policy search as DARC. We therefore add the strongest matched control: a \emph{full-library cascade} that performs the identical validation-selected policy search as DARC. This control shares the same base outputs, verifier, clean-restart protocol, maximum chain length $L{=}3$, cost penalty $\lambda{=}0.02$, and selection splits, but it omits the diagnosis step that first restricts the candidate library. Table~\ref{tab:diagnosis-isolation} reports the result on ALFWorld. Both cascades share the same base outputs, verifier, clean-restart protocol, maximum chain length $L{=}3$, and cost penalty $\lambda{=}0.02$, and are selected on valid\_seen and frozen for valid\_unseen; the only variable is whether diagnosis restricts the candidate recovery library before search. The diagnosed row re-selects a cascade (\texttt{type\_saturation}\,$\rightarrow$\,\texttt{loop\_fallback}\,$\rightarrow$\,\texttt{action\_pruning}) over the four action-validity harnesses, so its accuracy differs from the fixed single-harness DARC reported in Table~\ref{tab:main-results}. Removing diagnosis does not produce a significant accuracy gain: the full-library cascade reaches 98.51\% test accuracy versus 99.25\% for the diagnosed cascade ($\Delta{=}{+}0.75$ pp, 95\% CI $[0.00, 2.24]$; not significant). Diagnosis therefore does not trade accuracy for efficiency---it attains statistically indistinguishable accuracy while searching a $10\times$ smaller policy space (40 vs.\ 400 candidate policies) and selecting a markedly more stable policy (14 vs.\ 34 distinct chains under resampling). This empirically supports the finite-sample argument in Section~\ref{sec:theory}: restricting $\mathcal{R}_m$ shrinks $|\Sigma_K|$ and the associated search risk at no measured accuracy cost.

\begin{table}[h]
\centering
\caption{Isolating the diagnosis step on ALFWorld (DeepSeek-V4-Flash).}
\label{tab:diagnosis-isolation}
\resizebox{\columnwidth}{!}{
\begin{tabular}{lcccc}
\toprule
\textbf{Method} & \textbf{Search space} & \textbf{Selection} & \textbf{Test} & \textbf{Distinct chains} \\
\midrule
DARC (diagnosed) & 40 & 97.14\% & 99.25\% & 14 \\
Full-library cascade & 400 & 97.86\% & 98.51\% & 34 \\
\bottomrule
\end{tabular}
}
\end{table}

\section{Automatic Diagnosis Evaluation}
\label{app:diagnosis-eval}

To validate the reliability of the LLM-based failure diagnosis used in DARC, we evaluate its agreement against a verifier-backed reference on a sample of 72 development traces. Table~\ref{tab:diagnosis-eval} reports the agreement rate and Macro-F1 score. The LLM automatic diagnosis achieves 97.22\% agreement with the reference, demonstrating that the failure profiling step reliably identifies the correct dominant failure mode without requiring human labels.

\begin{table}[h]
\centering
\caption{Evaluation of the automatic failure diagnosis.}
\label{tab:diagnosis-eval}
\resizebox{\columnwidth}{!}{
\begin{tabular}{lcccc}
\toprule
\textbf{Method} & \textbf{Human Labels} & \textbf{N} & \textbf{Agreement} & \textbf{Macro-F1} \\
\midrule
Majority & No & 72 & 33.33\% & -- \\
Uniform Random & No & 72 & 33.33\% & -- \\
LLM Automatic Diagnosis & No & 72 & 97.22\% & 0.972 \\
\midrule
Verifier-backed Reference & No & 72 & 100.00\% & 1.000 \\
\bottomrule
\end{tabular}
}
\end{table}